\documentclass[11pt,a4paper]{article}
\usepackage[hyperref]{acl2020}
\usepackage{times}
\usepackage{latexsym}

\usepackage{times}
\usepackage{url}
\usepackage{latexsym}

\usepackage{longtable}
\usepackage{tabu}
\usepackage{arydshln}

\usepackage{graphicx}
\usepackage{wrapfig}
\usepackage{CJK,algorithm,algorithmic,amssymb,amsmath,array,epsfig,graphics,multirow,array,float,subfigure,verbatim,epstopdf}
\usepackage{enumitem}
\usepackage{color,soul}
\usepackage{hhline}
\usepackage{multirow}
\usepackage{xcolor}
\usepackage{hyperref}
\usepackage{tabularx}
\usepackage{bm}
\usepackage{seqsplit}
\usepackage{stmaryrd}
\usepackage{booktabs}
\usepackage{times}
\usepackage{latexsym}

\usepackage{microtype}

\aclfinalcopy 
\def\aclpaperid{3} 

\usepackage{xparse}

\title{ReviewRobot: Explainable Paper Review Generation based on Knowledge Synthesis}

\author{
Qingyun Wang$^{1}$, \ Qi Zeng$^1$, \ Lifu Huang$^{1}$, \\ \  \textbf{Kevin Knight}$^2$, \ \textbf{Heng Ji}$^{1}$, \  \textbf{Nazneen Fatema Rajani}$^3$\\
$^{1}$ University of Illinois at Urbana-Champaign $^{2}$ DiDi Labs $^{3}$ Salesforce Research\\
  \texttt{\fontfamily{pcr}\selectfont\{qingyun4,qizeng2,lifuh2,hengji\}@illinois.edu}\\
  {\tt kevinknight@didiglobal.com}\\
{\tt nazneen.rajani@salesforce.com}\\
}

\date{}

\begin{document}
\maketitle
\begin{abstract}

To assist human review process, we build a novel \emph{ReviewRobot} to automatically assign a review score and write  comments for multiple categories such as novelty and meaningful comparison. A good review needs to be \emph{knowledgeable}, namely that the comments should be constructive and informative to help improve the paper; and
\emph{explainable} by providing detailed evidence.
\emph{ReviewRobot} achieves these goals via three steps: (1) We perform domain-specific Information Extraction to construct a knowledge graph (KG) from the target paper under review, a related work KG from the papers cited by the target paper, and a background KG
from a large collection of previous papers in the domain. (2) By comparing these three KGs, we predict a review score
and detailed structured knowledge as evidence for each review category. (3)
We carefully select and generalize human review sentences into templates, and
apply these templates to transform the review scores and evidence into natural language comments.
Experimental results show that our review score predictor reaches 71.4\%-100\% accuracy.
Human assessment by domain experts shows that 41.7\%-70.5\% of the comments generated by \emph{ReviewRobot} are valid and constructive, and
better than human-written ones for 20\% of the time. Thus, ReviewRobot can serve as an assistant for paper reviewers, program chairs and authors.\footnote{The programs, data and resources are publicly available for research purpose at: \url{https://github.com/EagleW/ReviewRobot}}

\end{abstract}

\section{Introduction}

As the number of papers in our field increases exponentially, the reviewing practices today are more challenging than ever.
The quality of peer paper reviews is well-debated across the academic community~\cite{Bornmann2010,Mani2011,sculley2018avoiding,lipton2019troubling}.
How many times do we complain about a bad, random, dismissive, unfair, biased or inconsistent peer review? Authors even created various social groups at social media to release their frustrations and anger, such as the  ``Reviewer \#2 must be stopped'' group at Facebook\footnote{https://www.facebook.com/groups/reviewer2/}.
How many times are our papers rejected by a conference and then accepted by a better venue with only few changes?
As the number of
paper submissions continues to double or even triple every year, so does the need for high-quality peer reviews.

\begin{figure}
\centering
\includegraphics[width=\linewidth]{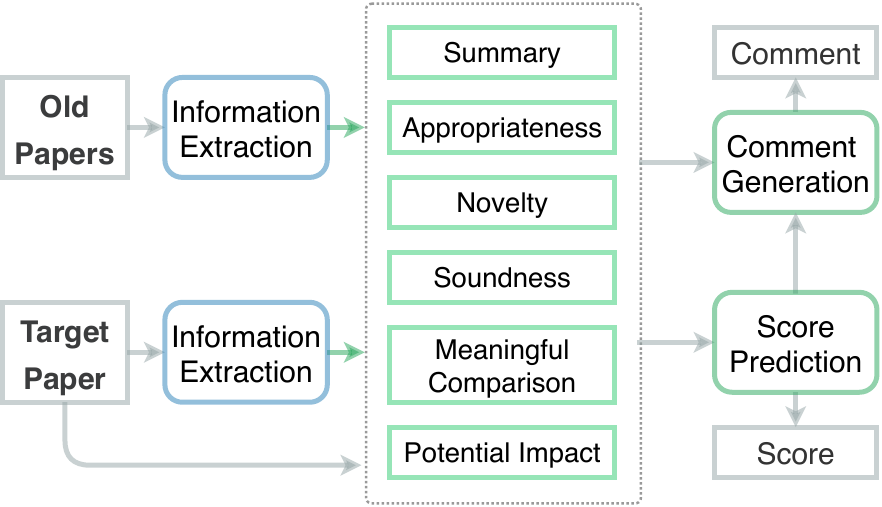}
\caption{\label{fig:overview}\emph{ReviewRobot} Architecture Overview}

\end{figure}
The following are two different reviews for the same paper rejected by ACL2019 and accepted by EMNLP2019 without any change on content:
\begin{itemize}
\item \textit{ACL 2019}:
Idea is too simple and tricky.
\item \textit{EMNLP 2019}: The main strengths of the paper lie in the interesting, relatively under-researched problem it covers, the novel and valid method and the experimental results.
\end{itemize}

\begin{figure*}[!hbt]
\centering
\includegraphics[width=\linewidth]{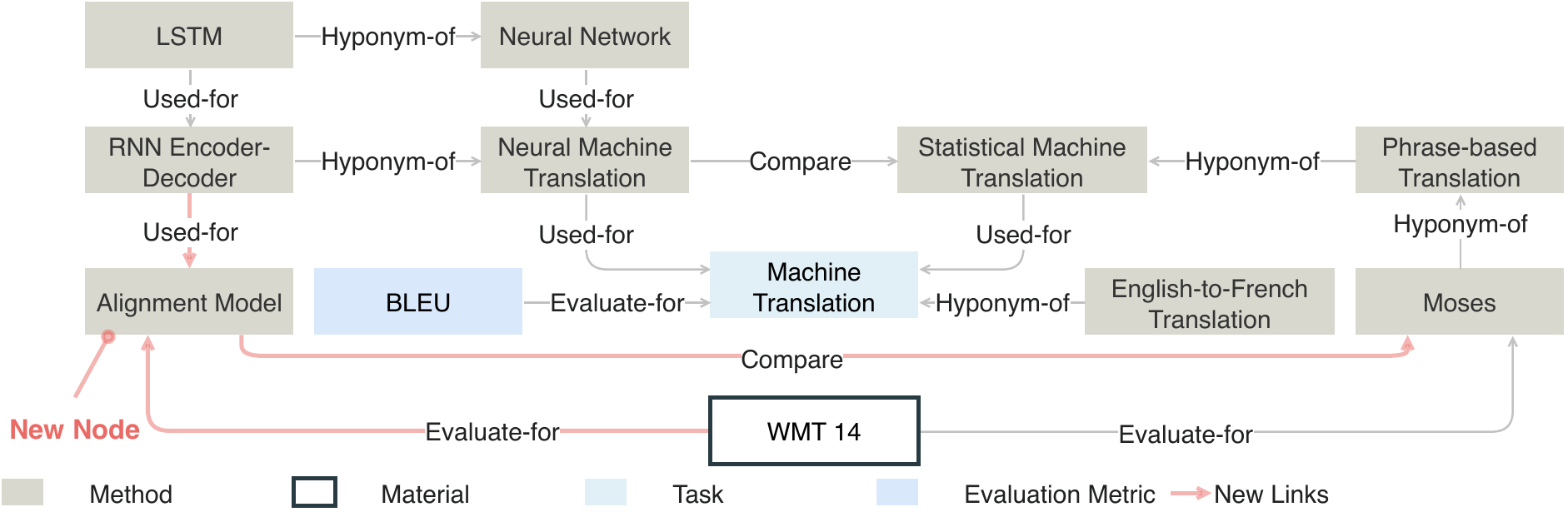}
\caption{Knowledge Graph Construction Example for Paper \protect\cite{atten15}}

\label{fig:kg}
\end{figure*}
These reviews, including the positive ones, are too vague and generic to be helpful.
We often see review comments stating a paper is missing references without pointing to any specific references, or criticizing an idea is not novel without showing similar ideas in previous work.
Some bad reviewers often ask to add citations to their own papers to inflate their citation record and h-index, and these papers are often irrelevant or published after the submission deadline of the target paper under review. Early study~\cite{Anderson2009} shows that the acceptance of a computer systems paper is often random and the dominant factor is the variability between reviewers. The inter-annotator agreement between two review scores for the ACL2017 accepted papers~\cite{kang-etal-2018-dataset} are only 71.5\%, 68.4\%, and 73.1\% for substance, clarity and overall recommendation respectively. \cite{Pier2018} found no agreement among reviewers in evaluating the same NIH grant application. The organizers of NIPS2014 assigned 10\% submissions to two different sets of reviewers and observed that these two committees disagreed for 25.9\% of the papers~\cite{Bornmann2010}, and half of NIPS2016 papers would have been rejected if reviews are done by a different group~\cite{Shah2017}.

These findings highlight the subjectivity in human reviews
and call for \textit{ReviewRobot}, an automatic review assistant to help human reviewers generate
knowledgeable and explainable review scores and comments, along with detailed evidence. We start by installing a brain for \emph{ReviewRobot} with a large-scale background knowledge graph (KG) constructed from previous papers in the target domain using domain-specific Information Extraction (IE) techniques. For each current paper under review, we apply the same IE method to construct two KGs, from its related work section and its other sections. By comparing the differences among these KGs, we extract pieces of evidence (e.g., novel knowledge subgraphs which are in the current paper but not in background KGs) for each review category and use them to predict review scores.
We manually select constructive human review sentences and generalize them into templates for each category. Then we apply these templates to convert
structured evidence to natural language comments for each category, using the predicted scores as a controlling factor.

Experimental results show that our review score predictor reaches 71.4\% overall accuracy on overall recommendation, which is very close to inter-human agreement (72.2\%).
The score predictor achieves 100\% accuracy for both appropriateness and impact categories.
Human assessment by domain experts shows that up to 70.5\% of the comments generated by ReviewRobot are valid, and
better than human-written ones 20\% of the time.

In summary, the major contributions of this paper are as follows:

\begin{itemize}
\item We propose a new research problem of generating paper reviews and present the first complete end-to-end framework to generate scores and comments for each review category.
\item Our framework is knowledge-driven, based on fine-grained
knowledge element comparison among papers, and thus the comments are highly explainable and constructive, supported by detailed evidence.
\item We create a new benchmark that includes 8K paper and review pairs, 473 manually selected pairs of paper sentences and constructive human review sentences, and a background KG constructed from 174K papers.
\end{itemize}






\section{Approach}

\subsection{Overview}

\begin{table*}[!tbh]
\vspace{-0.3cm}
\centering
\small
\setlength\tabcolsep{3pt}
\setlength\extrarowheight{1pt}
\setlist[itemize]{leftmargin=11pt, rightmargin=-60pt,before=\vspace*{-.6em},after=\vspace*{-.8em}}
\begin{tabularx}{\linewidth}{>{\hsize=0.45\hsize}X>{\hsize=1.55\hsize}X>{\hsize=1\hsize}X}
\toprule
\textbf{Category} & \multicolumn{1}{c}{\textbf{Evidence}} & \multicolumn{1}{c}{\textbf{Example}}\\
\midrule
Summary
&
\begin{itemize}[nosep]
    \item
$G_{P_\tau}$
\end{itemize}
&\begin{minipage}{\linewidth}
\includegraphics[width=1.4\linewidth]{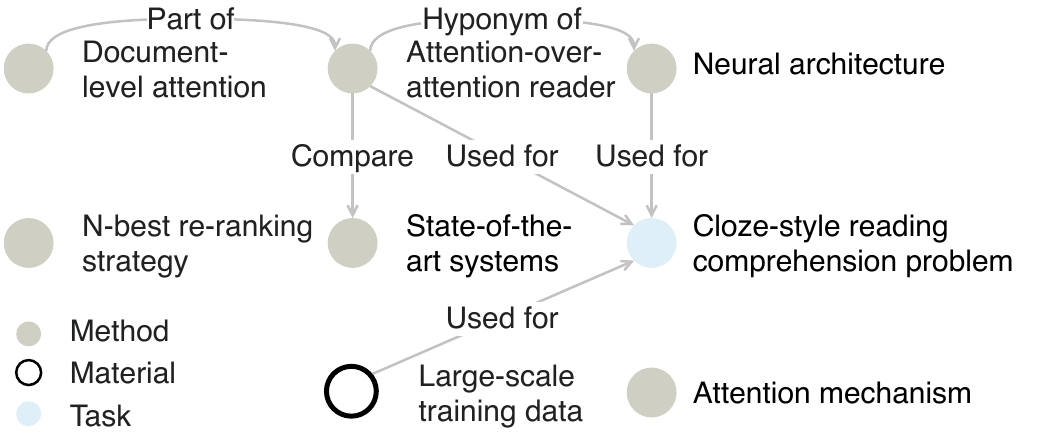}
\end{minipage}
\\\hdashline
Appropriateness
&
\begin{itemize}[nosep]
    \item The number of entities
    overlapped between
    the target paper
    and the domain's background KG:
    $|\{v|v\in G_{P_\tau} \cap G_B\}|$
    \item Abstract
\end{itemize}
&
\vspace{.01em}
\begin{minipage}{\linewidth}
\includegraphics[width=\linewidth]{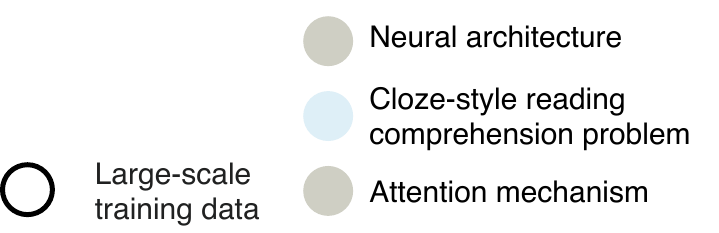}
\end{minipage}
\\\hdashline
Novelty
&
\begin{itemize}[nosep]
    \item New knowledge elements that appear in the target paper but not in the background KG:
    $|G_{P_\tau}- G_B|$
    \item Paper sentences that contain new knowledge elements
\end{itemize}
&\vspace{.01em}\begin{minipage}{\linewidth}
\includegraphics[width=1.4\linewidth]{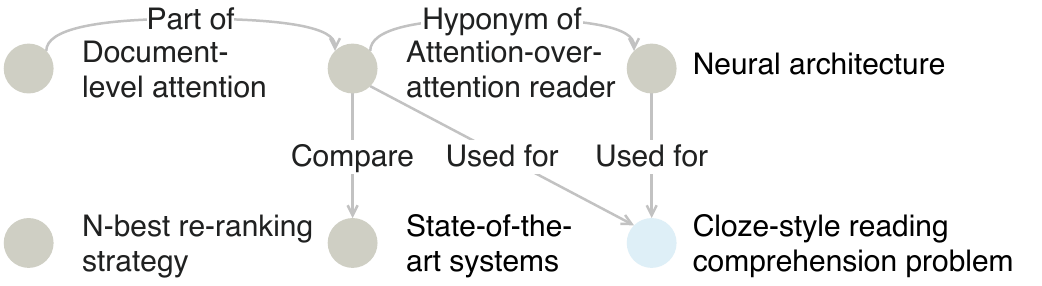}
\end{minipage}
\\\hdashline
Soundness
&
\begin{itemize}[nosep]
    \item The number of knowledge elements that appear in the contribution claims in the introduction section and that are verified in the experiment section
    \item Abstract
\end{itemize}
&
\begin{itemize}[nosep]
\item  attention-over-attention reader, n-best re-ranking strategy is verified in the related work section
\end{itemize}
\\\hdashline
Meaningful Comparison
&
\begin{itemize}[nosep]
    \item The number of papers about relevant knowledge elements which are missed in the related work section: $(G_B\cap G_{P_\tau}) - \bar{G}_{P_\tau}$
    \item The number of papers about relevant knowledge elements which are claimed new in the related work section: $ G_B\cap G_{P_\tau}\cap \bar{G}_{P_\tau}$
    \item The description sentences about comparison with related work
    \item If the related work section is not available, we use the difference between $G_{P_\tau}$and $G_B$ instead
\end{itemize}
&
\vspace{.01em}\begin{minipage}{\linewidth}
\includegraphics[width=\linewidth]{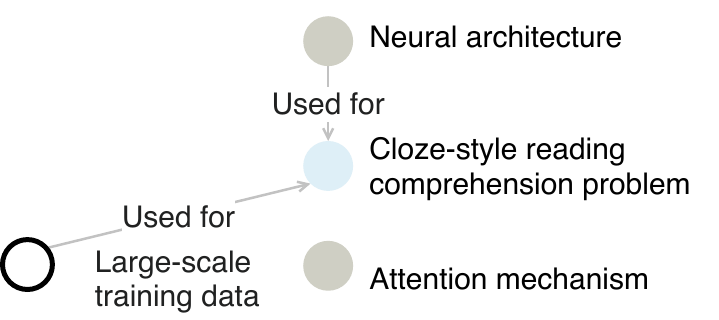}
\end{minipage}

\protect\cite{atten15,hermann2015teaching}
\\\hdashline
Potential Impact
&
\begin{itemize}[nosep]
    \item The number of new knowledge elements in the future work section
    \item The number of new software, systems, data sets, and other resources
\end{itemize}
&
\begin{itemize}[nosep]
    \item 5 new knowledge elements
    \item 1 new architecture
\end{itemize}
\\\hdashline
Overall Recommendations
&
\begin{itemize}[nosep]
    \item All features mentioned in the above categories
    \item Abstract
\end{itemize}
&
\\
\bottomrule
\end{tabularx}
\caption{Evidence Extraction for the example paper \emph{Attention-over-Attention Neural Networks for Reading Comprehension
} \protect\cite{cui-etal-2017-attention} 
}
\label{table:evidence}
\end{table*}

Figure~\ref{fig:overview} illustrates the overall architecture of \emph{ReviewRobot}. ReviewRobot first constructs knowledge graphs (KGs) for each target paper and a large collection of background papers, then it extracts evidence by comparing knowledge elements across multiple sections and papers, and uses the evidence to predict scores and generate comments for each review category.

We adopt the following most common categories from NeurIPS2019\footnote{\url{https://nips.cc/Conferences/2019/PaperInformation/ReviewerGuidelines}} and PeerRead~\cite{kang-etal-2018-dataset}:

\begin{table*}[!bht]
\centering
\small
\begin{tabularx}{\linewidth}{>{\hsize=0.35\hsize}X>{\centering\hsize=0.35\hsize}X>{\hsize=1.8\hsize}X>{\hsize=1.5\hsize}X}
\toprule
\textbf{Category} & \multicolumn{1}{c}{\textbf{\# of Pairs}} & \multicolumn{1}{c}{\textbf{Evidence  Sentence in Paper}}& \multicolumn{1}{c}{\textbf{Corresponding Review Sentence}}\\
\midrule
Summary & 236& In this paper, we present a simple but novel model called \textbf{attention-over-attention} reader for better solving cloze-style \textbf{reading comprehension} task. \protect\cite{cui-etal-2017-attention}
&The paper describes a new method called \textbf{attention-over-attention} for \textbf{reading comprehension}.\\\hdashline
Novelty & 33&The paper presents a new framework to solve the \textbf{SR} problem - amortized MAP inference and adopts a pre-learned affine projection layer to ensure the output is consistent with LR. \protect\cite{sonderby2016amortised}
&
It introduces a novel neural network architecture that performs a projection to the affine subspace of valid \textbf{SR} solutions ensuring that the high resolution output of the network is always consistent with the low resolution input.\\\hdashline
Soundness & 174& In high dimensions we empirically found that the \textbf{GAN} based approach, AffGAN produced the most visually appealing results. \protect\cite{sonderby2016amortised}
&Combined with \textbf{GAN}, this framework can obtain plausible and good results.\\\hdashline
Meaningful Comparison & 16& As a concrete instantiation, we show in this paper that we can enable recursive neural programs in the \textbf{NPI} model, and thus enable perfectly generalizable neural programs for tasks such as sorting where the original, \textbf{non-recursive NPI} program fails. \protect\cite{cai2017making}
&This paper improves significantly upon the original \textbf{NPI} work, showing that the model generalizes far better when trained on traces in recursive form.\\\hdashline
Potential Impact & 14& Since there may be several rounds of questioning and reasoning, these requirements bring the problem closer to task-oriented dialog and represent a significant increase in the difficulty of the challenge over the original \textbf{bAbI} (supporting fact) problems. \protect\cite{guo2016learning}
&I am a bit worried that the tasks may be too easy (as the \textbf{bAbI} tasks have been), but still, I think locally these will be useful.\\
\bottomrule
\end{tabularx}
\caption{Annotation Statistics and Examples for Template Generalization\label{table:annotation}\
}
\end{table*}

\begin{itemize}

\setlength{\itemsep}{0pt}
\setlength{\parsep}{0pt}
\setlength{\parskip}{0pt}

\item \textbf{Summary}: What is this paper about?
\item \textbf{Appropriateness}: Does the paper fit in the venue?
\item \textbf{Clarity}:
Is it clear what was done and why? Is the paper well-written and well-structured?
\item \textbf{Novelty}:
Does this paper break new ground in topic, methodology, or content? How exciting and innovative is the research it describes?
\item \textbf{Soundness}:
Can one trust the empirical claims of the paper -- are they supported by proper experiments and are the results of the experiments correctly interpreted?
\item \textbf{Meaningful Comparison}: Do the authors make clear where the problems and methods sit with respect to existing literature? Are the references adequate?
\item \textbf{Potential Impact}:  How significant is the work described? If the ideas are novel, will they also be useful or inspirational? Does the paper bring any new insights into the nature of the problem?

\end{itemize}

\subsection{Knowledge Graph Construction}

Generating meaningful and explainable reviews requires ReviewRobot to understand the knowledge elements of each paper.
We apply a state-of-the-art Information Extraction (IE) system for Natural Language Processing (NLP) and Machine Learning (ML) domains~\cite{luan-etal-2018-multi}
to construct the following knowledge graphs (KGs):

\begin{itemize}

\setlength{\itemsep}{0pt}
\setlength{\parsep}{0pt}
\setlength{\parskip}{0pt}

    \item $G_{P_\tau}$: A KG constructed from the abstract and conclusion sections of a target paper under review $P_\tau$, which describes the main techniques.

    \item $\bar{G}_{P_\tau}$: A KG constructed from the related work section of $P_\tau$,
    which describes related techniques.

    \item $G_B$: A background KG constructed from all of the old
    NLP/ML papers published before the publication year of $P_\tau$, in order to teach \emph{ReviewRobot} what's happening in the field.

\end{itemize}

Each node $v\in V$ in a KG represents an entity, namely a cluster of co-referential entity mentions, assigned one of six types: \emph{Task}, \emph{Method}, \emph{Evaluation Metric}, \emph{Material}, \emph{Other Scientific Terms}, and \emph{Generic Terms}.
Following the previous work on entity coreference for scientific domains \cite{koncel-kedziorski-etal-2019-text}, we choose the longest informative entity mention in each cluster to represent the entity. We consider two entity clusters from different papers as coreferential if one's representative mention appears in the other. Each edge represents a relation between two entities.  There are seven relation types: \emph{Used-for}, \emph{Feature-of}, \emph{Evaluate-for}, \emph{Hyponym-of}, \emph{Part-of}, \emph{Compare}, and \emph{Conjunction}.
Figure~\ref{fig:kg} shows an example KG constructed from~\cite{atten15}.

\subsection{Evidence Extraction}
We compare the differences among the constructed KGs to extract evidence for each review category. Table~\ref{table:evidence} shows the methods to extract evidence and some examples for each category.

\subsection{Score Prediction}

Following \cite{kang-etal-2018-dataset}, we consider review score prediction as a multi-label classification task.
For a target paper,
we first encode its category related sentences with an attentional Gated Recurrent Unit (GRU)~\cite{cho-etal-2014-learning,atten15}
to obtain attentional contextual sentence embedding. We also
encode the extracted evidence for each review category with an embedding layer.
Then we
concatenate the context embedding and evidence embedding
to predict the quality score $r$ in the range of 1 to 5
with a linear output layer.
We use the prediction probability as the confidence score.

\begin{table*}[ht!]
\small
\begin{tabularx}{\linewidth}{>{\centering\hsize=1.16\hsize}X>{\centering\arraybackslash\hsize=0.98\hsize}X>{\centering\arraybackslash\hsize=0.98\hsize}X>{\centering\arraybackslash\hsize=0.98\hsize}X>{\centering\arraybackslash\hsize=0.98\hsize}X>{\centering\arraybackslash\hsize=0.98\hsize}X>{\centering\arraybackslash\hsize=0.98\hsize}X>{\centering\arraybackslash\hsize=0.98\hsize}X>{\centering\arraybackslash\hsize=0.98\hsize}X}
\toprule
\multirow{2}{*}{\textbf{Conference}}&\multicolumn{8}{c}{\textbf{Year}}\\  \cline{2-9}
&\textbf{2013}&\textbf{2014}&\textbf{2015}&\textbf{2016}&\textbf{2017}&\textbf{2018}&\textbf{2019}&\textbf{2020}\\
\midrule
ICLR&-&-&-&-&404&874&1,342&2,067\\\hdashline
NeurIPS&342&399&389&545&655&963&-&-\\\hdashline
ACL&-&-&-&-&130&-&-&-\\
\bottomrule
\end{tabularx}
\caption{\label{table:dataset} Data Statistics for Paper Review Corpus (\# of papers)
}
\end{table*}

\begin{table*}[ht!]
\small
\setlength\tabcolsep{2pt}
\begin{tabularx}{\linewidth}{>{\centering\hsize=1.54\hsize}X>{\centering\arraybackslash\hsize=0.94\hsize}X>{\centering\arraybackslash\hsize=0.94\hsize}X>{\centering\arraybackslash\hsize=0.94\hsize}X>{\centering\arraybackslash\hsize=0.94\hsize}X>{\centering\arraybackslash\hsize=0.94\hsize}X>{\centering\arraybackslash\hsize=0.94\hsize}X>{\centering\arraybackslash\hsize=0.94\hsize}X>{\centering\arraybackslash\hsize=0.94\hsize}X>{\centering\arraybackslash\hsize=0.94\hsize}X}\toprule
\textbf{Years (1965$\sim$)}&\textbf{2011}&\textbf{2012}&\textbf{2013}&\textbf{2014}&\textbf{2015}&\textbf{2016}&\textbf{2017}&\textbf{2018}&\textbf{2019}\\
\midrule
\# of Entities&535,075& 585,321& 628,713& 683,686& 737,878& 801,740& 870,992& 950,457& 1,008,955\\\hdashline
\# of Relations&160,123& 175,780& 188,876& 205,898& 222,592& 242,312& 263,827& 288,805& 307,636\\
\bottomrule
\end{tabularx}
\caption{\label{table:bgkb} Data Statistics for Background Knowledge Graphs since 1965}
\end{table*}

\begin{table*}[!htb]
\centering
\small
\begin{tabularx}{\linewidth}{>{\arraybackslash\hsize=2.1\hsize}X>{\centering\arraybackslash\hsize=1\hsize}X>{\centering\arraybackslash\hsize=2.3\hsize}X>{\centering\arraybackslash\hsize=0.5\hsize}X>{\centering\arraybackslash\hsize=1.1\hsize}X>{\centering\arraybackslash\hsize=0.6\hsize}X>{\centering\arraybackslash\hsize=0.5\hsize}X>{\centering\arraybackslash\hsize=1.1\hsize}X>{\centering\arraybackslash\hsize=0.6\hsize}X>{\centering\arraybackslash\hsize=0.5\hsize}X>{\centering\arraybackslash\hsize=1.1\hsize}X>{\centering\arraybackslash\hsize=0.6\hsize}X}
\toprule
 \multirow{2}{*}{\textbf{Category}}      & \multirow{2}{*}{\begin{tabular}[c]{@{}c@{}}\textbf{Human} \\\textbf{Kappa} \\\textbf{Score}\end{tabular}} & \multirow{2}{*}{\begin{tabular}[c]{@{}c@{}}\textbf{Human Average}\\\textbf{Inter  Annotator}\\\textbf{Agreement}\end{tabular}} & \multicolumn{3}{c}{\textbf{CNN} \cite{kang-etal-2018-dataset}}                    & \multicolumn{3}{c}{\begin{tabular}[c]{@{}c@{}}\textbf{GRU with Abstract}\end{tabular}}      & \multicolumn{3}{c}{\begin{tabular}[c]{@{}c@{}}\textbf{GRU with Evidence}\end{tabular}}    \\ \cline{4-12}
                             &           &                       & \begin{tabular}[c]{@{}c@{}}\textbf{Score} \\\textbf{Acc.} \end{tabular}& \begin{tabular}[c]{@{}c@{}} \textbf{Decision} \\\textbf{Acc.}\end{tabular}  & \textbf{MSE} & \begin{tabular}[c]{@{}c@{}} \textbf{Score} \\\textbf{Acc.}\end{tabular} & \begin{tabular}[c]{@{}c@{}} \textbf{Decision} \\\textbf{Acc.}\end{tabular}  & \textbf{MSE} & \begin{tabular}[c]{@{}c@{}} \textbf{Score} \\\textbf{Acc.}\end{tabular} & \begin{tabular}[c]{@{}c@{}} \textbf{Decision} \\\textbf{Acc.}\end{tabular} & \textbf{MSE}\\ 
\midrule
 Recommendation               & 33.63    & 72.2               &{71.43}                &57.14    & 0.714 &{71.43}&57.14      &  0.714  & \textbf{85.71}                & \textbf{71.43}                                    &   \textbf{0.571}        \\ \hdashline
Appropriateness                & 100     & 100                &\textbf{85.71}                & \textbf{100}    &  \textbf{0.143}   &\textbf{85.71}                & \textbf{100}    &  \textbf{0.143}   & \textbf{85.71}                & \textbf{100}                        &                 \textbf{0.143}         \\ \hdashline
\begin{tabular}[c]{@{}l@{}}Meaningful \\Comparison\end{tabular}                      & 100     & 100                &57.14                &57.14    &  0.857 &57.14&  71.42      &  0.857 & \textbf{71.43}                & \textbf{71.43}                   &                           \textbf{0.714}     \\ \hdashline
Soundness                      & 100     & 100                &42.86                &42.86      &  1.86   &14.28&    \textbf{85.71}  & 0.857 & \textbf{71.43}                & \textbf{85.71} &                                      \textbf{0.714}          \\\hdashline
Novelty                    & 100     & 100                &42.86                &42.86     &  2.29      &28.57                &28.57     & 2.43& \textbf{71.43}                & \textbf{71.43}                       &                \textbf{0.714}         \\ \hdashline
Clarity                        & 70.20   & 86.11              &\textbf{42.86}                &\textbf{71.43}  & \textbf{1.00} &\textbf{42.86}                & \textbf{71.43}     &      \textbf{1.00}    & \textbf{42.86}                & \textbf{71.43}               &                     \textbf{1.00}               \\\hdashline
\begin{tabular}[c]{@{}l@{}}Potential\\ Impact     \end{tabular}                    & 100     & 100                &\textbf{85.71}                &\textbf{100} & \textbf{0.143} &\textbf{85.71}                &\textbf{100}             & 0.571  & \textbf{85.71}                & \textbf{100}                   &                 \textbf{0.143}            \\
\bottomrule
\end{tabularx}
\caption{Score Prediction Accuracy (\%) and Mean Square Error (MSE) on ACL2017 Data Set  \label{table:score prediction}
}
\end{table*}

\subsection{Comment Generation}

Given the evidence graphs and predicted scores as input, we perform template-based comment generation for each category. We aim to learn good templates from human reviews. Unfortunately as we have discussed earlier, not all human written review sentences are of high quality, even for those accepted papers. Therefore in order to generalize templates, we need to carefully select those constructive and informative human review sentences that are supported by certain evidence in the papers. To avoid expensive manual selection, we design a semi-automatic bootstrapping approach.
We manually annotate 200 paper-review pairs from ACL2017 and ICIR2017 datasets, and then use them as seed annotations to train an attentional GRU \cite{cho-etal-2014-learning} based binary (select/not select) classifier to process the remaining human review sentences and keep high-quality reviews with high confidence. Our attentional GRU achieves binary classification accuracy 85.25\%.
Table \ref{table:annotation} shows the annotation statistics and some examples.

For appropriateness, soundness, and potential impact categories, we generate generic positive or negative comments based on the predicted scores. For summary, novelty, and meaningful comparison categories, we consider review generation as a template-based graph-to-text generation task. Specifically, for summary and novelty, we generate reviews by describing the \emph{Used-for}, \emph{Feature-of}, \emph{Compare} and \emph{Evaluate-for} relations in evidence graphs. We choose positive or negative templates depending on whether the predicted scores are above 3.
We use the predicted overall recommendation score to control summary generation.
For related work, we keep the knowledge elements in the evidence graph with a TF-IDF score~\cite{jones1972statistical} higher than 0.5. For each knowledge element, we recommend the most recent 5 papers that are not cited as related papers.

\section{Experiments}

\subsection{Data}
We choose papers in NLP and ML domains in our experiments because it's easy for us to analyze results, and we are not the most harsh community in Computer Science: the average review score in our corpus is 3.3 out of 5 while it is 2.5/5 in the computer system area~\cite{Anderson2009}. In addition to the review corpus constructed by~\cite{kang-etal-2018-dataset}, we have collected additional paper-review pairs from openreview\footnote{ We collect ICLR paper using open review API \url{https://openreview-py.readthedocs.io/}} and NeurIPS\footnote{\url{https://papers.nips.cc/}}. In total, we have collected 8,110 paper and review pairs as shown in Table~\ref{table:dataset}. We construct the background KG from 174,165 papers from the open research corpus~\cite{ammar-etal-2018-construction}.
Table~\ref{table:bgkb} shows the data statistics of background KGs.

\begin{figure}[hbt!]
\centering
\includegraphics[width=\linewidth]{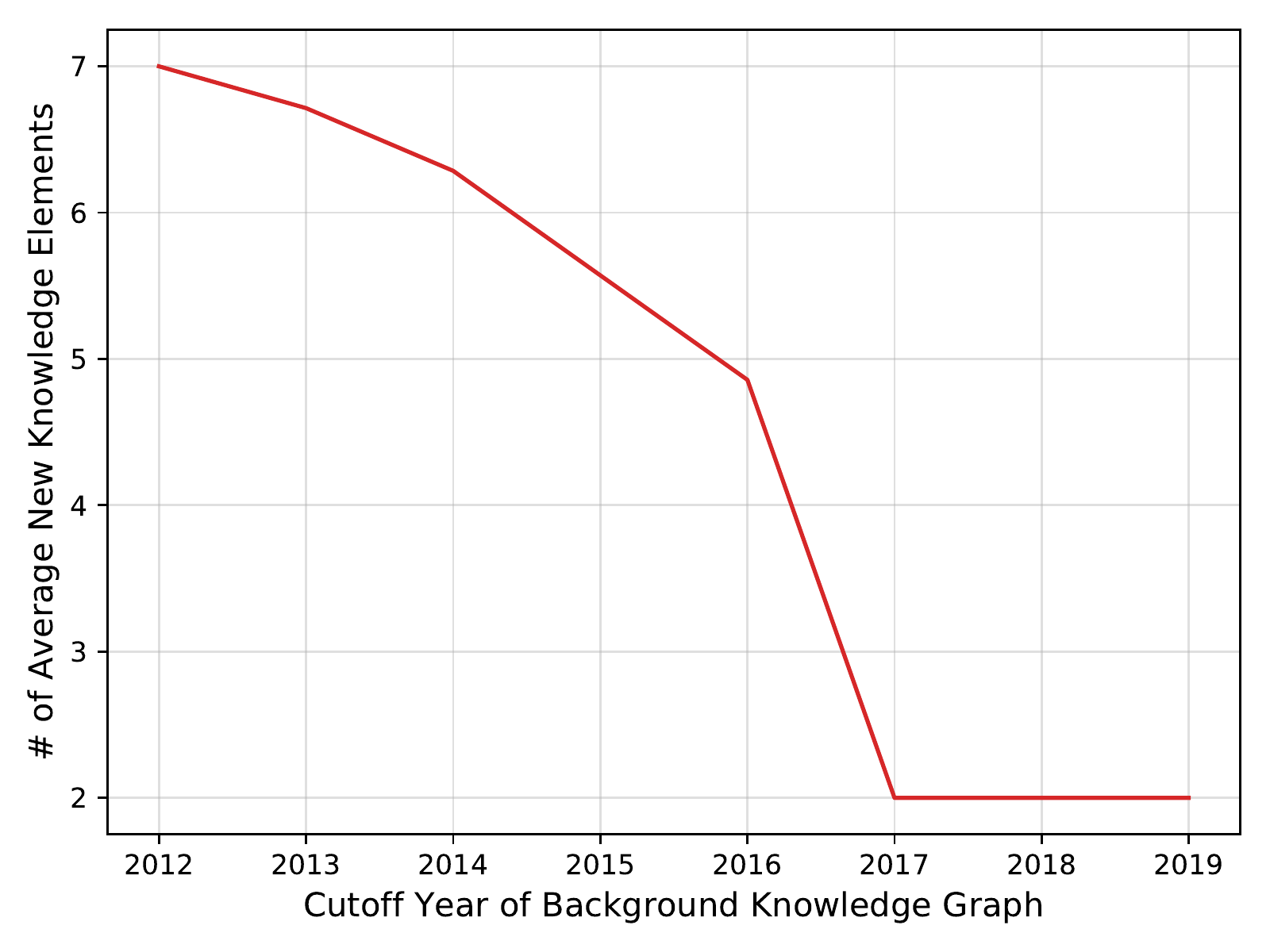}
\caption{\label{fig:novelty} The average number of new knowledge elements in ACL2017 test papers given the background KG constructed from (1965$\sim$cutoff year)}
\end{figure}
\subsection{Score Prediction Performance}

We use the ACL2017 dataset in the score prediction task because it has complete score annotations for each review category.
We follow the data split of PeerRead~\cite{kang-etal-2018-dataset}\footnote{We exclude the training pairs that we fail to run IE system on. The test set remains the same as \cite{kang-etal-2018-dataset}.}.
Unlike PeerRead which uses multiple review scores for the same input paper, we use the rounded average score of each category as the target score. Table~\ref{table:score prediction} shows that our model trained from carefully selected constructed reviews has already reached a prediction accuracy of 71.43\% for overall recommendation, which is very close to the human inter-annotator agreement (72.2\%) and dramatically advances state-of-the-art approaches in most categories. Our model also produces the lowest  mean square errors for all categories.

Our knowledge graph synthesis based approach is particularly effective at predicting Novelty score and achieves the accuracy of 71.43\%, which is much higher than the accuracy (28.57\%) of all other automatic prediction methods using paper abstracts only as input. In Figure~\ref{fig:novelty} we show the average number of new knowledge elements of our test set consisting of ACL2017 papers, when it's reviewed during different years. When the background KG includes newer work, the novelty of these papers decreases, especially after 2017. This indicates that our approach provides a reliable measure for computing novelty.

As a fun experiment, we also run \emph{ReviewRobot} on this paper submission itself. The predicted review scores are 5, 3, 4, 3, 4, 4, and 4 for Appropriateness, Meaningful Comparison, Soundness, Novelty, Clarity, Potential Impact and Overall Recommendation, respectively, which means this paper is likely to be accepted.

\subsection{Comment Generation Performance}

For the system generated review comments for 50 ACL2017 papers, we ask domain experts to check whether each comment is \emph{constructive} and \emph{valid}.
Two researchers independently annotate the reviews and reach the inter-annotator agreement of 92\%, 92\%, and 82\% for Novelty, Summary and Related Work, respectively. One expert annotator performs data adjudication.
The percentages of constructive and valid comments are 70.5\%, 44.6\% and 41.7\% for Summary, Novelty and Meaningful Comparison, respectively. Human assessors also find that for 20\% of these papers, human reviewers do not suggest missing related work for Meaningful Comparison, while \emph{ReviewRobot} generates constructive and informative comments. For example, the human reviewer states ``The paper would be stronger with the inclusion of more baselines based on related work\footnote{Review for \cite{niu-etal-2017-improved} \url{https://github.com/allenai/PeerRead/blob/master/data/acl_2017/train/reviews/318.json}}'', but fails to provide any useful references.
In the following we compare the human and system generated reviews for an example paper~\cite{niu-etal-2017-improved}:

\noindent\textbf{Summary}

*  {[SYSTEM] \textit{The paper proposes novel skip-gram, attention scheme, sememe-encoded models and word representation learning for NLP tasks. The authors uses linguistic common-sense knowledge bases. }}

*  {[HUMAN] \textit{This paper proposes the use of HowNet to enrich embeddings. }}

\noindent\textbf{Novelty}

* {[SYSTEM] \textit{The paper proposes novel attention scheme and sememe-encoded models.  }}

*  {[HUMAN] \textit{I see the main contribution of the work to be the results which show that we can learn better word representations (unsure about wsi) by modeling sememe information than other competitive baselines.}}

\noindent\textbf{Meaningful Comparison}

* {[SYSTEM]\textit{ The following related papers are missing:
    \begin{enumerate}
\item  About low-dimensional semantic space:
 \begin{enumerate}
	\item Unsupervised approximate-semantic vocabulary learning for human action and video classification \cite{zhao2013unsupervised}
Qiong Zhao and Horace HS Ip. 2013. Unsupervised Approximate-semantic Vocabulary Learning for Human Action and Video Classification. Pattern Recognition Letters, 34(15):1870–1878.
    \end{enumerate}
\item About sememes:
 \begin{enumerate}
	\item
	Chinese Word Sense Disambiguation with PageRank and HowNet \cite{wang-etal-2008-chinese}:
    Jinghua Wang, Jianyi Liu, and Ping Zhang. 2008. Chinese Word Sense Disambiguation with PageRank and HowNet. In Proceedings of the Sixth SIGHAN Workshop on Chinese Language Processing.
	\item
	A maximum entropy approach to HowNet-based Chinese word sense disambiguation \cite{wong-yang-2002-maximum}:
    Ping Wai Wong and Yongsheng Yang. 2002. A Maximum Entropy Approach to HowNet-based Chinese
    Word Sense Disambiguation. In COLING-02: SEMANET: Building and Using Semantic Networks.
    \end{enumerate}
\item About word similarity and word analogy:
 \begin{enumerate}
	\item Open IE as an Intermediate Structure for Semantic Tasks \cite{stanovsky-etal-2015-open}:
Gabriel Stanovsky, Ido Dagan, et al. 2015. Open IE as an Intermediate Structure for Semantic Tasks. In Proceedings of the 53rd Annual Meeting of the Association for Computational Linguistics and the 7th International Joint Conference on Natural Language Processing (Volume 2: Short Papers), pages 303–308.
    \end{enumerate}
    \end{enumerate}
}}

* {[HUMAN]  \textit{The paper would be stronger with the inclusion of more baselines based on related work.}}

\subsection{Remaining Challenges and Limitations}

The quality of \emph{ReviewRobot} is mainly limited by state-of-the-art Information Extraction performance for the scientific literature domain. In the future we plan to annotate more data to cover  more dimensions for paper profiling (such as goal and main contribution), and
more fine-grained
knowledge types to improve the extraction quality. For example, for the NLP domain we can extract finer-grained subtypes: a model can include parameters, components and features. The goal of an NLP paper could belong to: ``\textit{New methods for specific NLP problems}'', ``\textit{End-user applications}'', ``\textit{Corpora and evaluations}'', ``\textit{New machine learning methods for NLP}'', ``\textit{Linguistic theories }'', ``\textit{Cognitive modeling and psycholinguistic research}'' or ``\textit{Applications to social sciences and humanities}''. Our current evidence extraction framework also lacks of a salience measure to assign different weights to different types of knowledge elements.

Paper review generation requires background knowledge acquisition and comparison with the target paper content.
Our novel approach on constructing background KG has helped improve the quality of review comments on novelty but the KG is still too flat to generate comments on soundness.
For example, from the following two sentences in a paper: ``Third, at least 93\% of time expressions contain at least one time token.'', and ``For the relaxed match on all three datasets , SynTime-I and SynTime-E achieve recalls above 92\%.'', a knowledgeable human reviewer can infer 93\% as the upper bound of performance and write a comment: ``Section 5.2 : given this approach is close to the ceiling of performance since 93 \% expressions contain time token , and the system has achieved 92 \% recall , how do you plan to improve further?''. Similarly, \emph{ReviewRobot} cannot generalize knowledge elements into high-level comments such as ``\textit{deterministic}'' as in ``\textit{The tasks 1-5 are also completely deterministic}''.

\emph{ReviewRobot} still lacks of deep knowledge reasoning ability to judge the soundness of algorithm design details, such as whether the split of data set makes sense, whether a model is able to generalize.
\emph{ReviewRobot} is not able to comment on missing hypotheses, the problems on experimental setting and future work. \emph{ReviewRobot} currently focuses on text only and cannot comment on mathematical formulas, tables and figures.

Good machine learning models rely on good data. We need massive amounts of good human reviews to fuel \emph{ReviewRobot}. In our current approach, we manually select a subset of good human review sentences that are also supported by corresponding sentences in the target papers. This process is very time-consuming and expensive. We need to build a better review infrastructure in our community, e.g., asking authors to provide feedback and rating to select constructive reviews as in NAACL2018\footnote{https://naacl2018.wordpress.com/2018/02/26/acceptance-and-author-feedback/}.

\section{Related Work}

\noindent \textbf{Paper Acceptance Prediction.} \citet{kang-etal-2018-dataset} has constructed a paper review corpus, PeerRead, and trained paper acceptance classifiers.
\citet{Huang2018} applies an interesting visual feature to compare the pdf layouts and proves its effectiveness to make paper acceptance decision.
\citet{ghosal-etal-2019-deepsentipeer} applies sentiment analysis features to improve acceptance prediction. The KDD2014 PC chairs exploit author status and review comments for predicting paper acceptance~\cite{Leskovec2014}.
We extend these methods to score prediction and comment generation with detailed knowledge element level evidence for each specific review category.

\noindent \textbf{Paper Review Generation.}
\citet{Bartoli2016} proposes the first deep neural network framework to generate paper review comments.
The generator is trained with 48 papers from their own lab.
In comparison, we perform more concrete and explainable review generation by predicting scores and generating comments for each review category following a rich set of evidence, and use a much larger data set.
\citet{nagata-2019-toward} generates comment sentences to explain grammatical errors as feedback to improve paper writing. \cite{xing-etal-2020-automatic, Luu2020CitationTG} extract paper-paper relations and use them to guide citation text generation.

\noindent \textbf{Review Generation in other Domains.}
Automatic review generation techniques have been applied to many other domains including music~\cite{tata-di-eugenio-2010-generating},
restaurants~\cite{oraby-etal-2017-harvesting,juuti2018stay,li-etal-2019-generating,brazinskas-etal-2020-unsupervised},
and products~\cite{catherine2018transnets,li-etal-2019-generating,li-tuzhilin-2019-towards,dong-etal-2017-learning-generate,ni-mcauley-2018-personalized,brazinskas-etal-2020-unsupervised}.
These methods generally apply a sequence-to-sequence model with attention to aspects and attributes (e.g. food type).
Compared to these domains, paper review generation is much more challenging because it requires the model to perform deep understanding on paper content, construct knowledge graphs to compare knowledge elements across sections and papers, and synthesize information as input evidence for comment generation.

\noindent \textbf{Controlled Knowledge-Driven Generation.}
There have been some other studies on text generation controlled by sentiment~\cite{hu2017toward}, topic~\cite{krishna-srinivasan-2018-generating},
text style~\cite{shen2017style,liu-etal-2019-towards-comprehensive,tikhonov-etal-2019-style},
and facts~\cite{wang-etal-2020-towards}.
The usage of external supportive knowledge in text generation can be roughly divided into the following three levels: (1) Knowledge Description, which transforms structured data into unstructured text, such as Table-to-Text Generation~\cite{mei-etal-2016-talk,lebret-etal-2016-neural,chisholm-etal-2017-learning,sha2017order,DBLP:conf/aaai/LiuWSCS18,nema-etal-2018-generating,wang-etal-2018-describing,moryossef-etal-2019-step,nie-etal-2019-encoder,castro-ferreira-etal-2019-neural,wang-etal-2020-towards,shahidi-etal-2020-two} %
and its variants in low-resource ~\cite{ma-etal-2019-key} and multi-lingual setting~\cite{kaffee-etal-2018-learning,kaffee2018mind}, Data-to-Document \cite{wiseman-etal-2017-challenges,puduppully-etal-2019-data,gong-etal-2019-enhanced,iso-etal-2019-learning}, Graph-to-Text~\cite{flanigan-etal-2016-generation,song-etal-2018-graph,zhu-etal-2019-modeling,koncel-kedziorski-etal-2019-text}, and Topic-to-text~\cite{tang-etal-2019-topic}, and
Knowledge Base Description~\cite{kiddon-etal-2016-globally,gardent-etal-2017-creating,koncel-kedziorski-etal-2019-text};
(2) Knowledge Synthesis, which retrieves knowledge base and organizes text answers, such as Video Caption
Generation~\cite{whitehead-etal-2018-incorporating}, KB-supported Dialogue Generation~\cite{han-etal-2015-exploiting,zhou2018commonsense,parthasarathi-pineau-2018-extending,liu-etal-2018-knowledge,young2018augmenting,wen-etal-2018-sequence,chen-etal-2019-enhancing,liu-etal-2019-knowledge}, Knowledge-guided comment Generation~\cite{li-etal-2019-coherent}, paper generation~\cite{wang-etal-2018-paper,wang-etal-2019-paperrobot,Cachola2020TLDRES}
, and abstractive summarization~\cite{gu-etal-2016-incorporating,sharma-etal-2019-entity,huang-etal-2020-knowledge}.

\section{Application Limitations and Ethical Statement}
The types of evidence we have designed in this paper are limited to NLP, ML or related areas, and thus they are not applicable to other scientific domains such as biomedical science and chemistry.
Whether \emph{ReviewRobot} is essentially beneficial to the scientific community also depends on who uses it. Here are some example scenarios where \emph{ReviewRobot} should and should not be used:

\begin{itemize}
\item \textbf{Should-Do}: Reviewers use \emph{ReviewRobot} merely as an assistant to write more constructive comments and compare notes.
\item \textbf{Should-Do}: Editors use \emph{ReviewRobot} to assist filtering very bad papers during screening.
\item \textbf{Should-Do}: Authors use \emph{ReviewRobot} to get initial feedback to improve paper writing such as adding missing references and highlighting the recommended novel points.
\item \textbf{Should-Do}: Researchers use \emph{ReviewRobot} to perform literature survey, find more good papers and validate the novelty of their papers.
\item \textbf{Should-Not-Do}: Reviewers submit \emph{ReviewRobot}'s output without reading the paper carefully.
\item \textbf{Should-Not-Do}: Editors send \emph{ReviewRobot}'s output and make decisions based on it.
\item \textbf{Should-Not-Do}: Authors revise their papers to fit into \emph{ReviewRobot}'s features to boost review scores. For example, authors should not deliberately cite all related papers or add irrelevant new terms to boost their review scores.
\end{itemize}

\section{Conclusions and Future Work}
We build a \emph{ReviewRobot} for predicting review scores and generating detailed comments for each review category, which can serve as an effective assistant for human reviewers and authors who want to polish their papers. The key innovation of our approach is to construct knowledge graphs from the target paper and a large collection of in-domain background papers, and summarize the pros and cons of each paper on knowledge element level with detailed evidence. We plan to enhance \emph{ReviewRobot}'s knowledge reasoning capability by building a taxonomy on top of the background KG, and incorporating multi-modal analysis of formulas, tables, figures, and citation networks.

\section*{Acknowledgments}
The knowledge extraction and prediction components were supported by the U.S. NSF No. 1741634 and Air Force No. FA8650-17-C-7715. The views and conclusions contained in this document are those of the authors and should not be interpreted as representing the official policies, either expressed or implied, of the U.S. Government. The U.S. Government is authorized to reproduce and distribute reprints for Government purposes notwithstanding any copyright notation here on.

\bibliography{anthology,acl2020}
\bibliographystyle{acl_natbib}

\end{document}